\documentclass[letterpaper, 10 pt, journal, twoside]{IEEEtran}
\usepackage{amsmath,amsfonts}
\usepackage{algorithmic}
\usepackage{algorithm}
\usepackage{array}
\usepackage[caption=false,font=normalsize,labelfont=sf,textfont=sf]{subfig}
\usepackage{textcomp}
\usepackage{stfloats}
\usepackage{url}
\usepackage{verbatim}
\usepackage{graphicx}
\usepackage{cite}
\usepackage{color}
\usepackage{xcolor}
\colorlet{BLACK}{black}
\newcommand{\shan}[1]{\textcolor{black}{#1}}
\newcommand{\zc}[1]{\textcolor{black}{#1}}
\newcommand{\green}[1]{\textcolor{black}{#1}}
\newcommand{\red}[1]{\textcolor{black}{#1}}
\newcommand{\blue}[1]{\textcolor{black}{#1}}

\newcommand{\final}[1]{\textcolor{black}{#1}}
\newcommand{\finall}[1]{\textcolor{black}{#1}}

\begin{document}

\title{Tacchi: A Pluggable \zc{and Low Computational Cost} Elastomer Deformation Simulator for Optical Tactile Sensors}

\author{Zixi Chen \final{$^{1,2}$}, Shixin Zhang\final{$^{3}$}, Shan Luo\final{$^{1}$}*, Fuchun Sun\final{$^{2}$}, and Bin Fang\final{$^{2}$}* 

 \thanks{\final{Manuscript received August 16, 2022; Revised October 31, 2022; Accepted January 11, 2023.This paper was recommended for publication by Editor A. Bera upon evaluation of the Associate Editor and Reviewers’ comments.This work was  jointly supported by Major Project of the New Generation of Artificial Intelligence, China (No. 2018AAA0102900), the National Natural Science Foundation of China (Grant No. 62173197) and the EPSRC project "ViTac: Visual-Tactile Synergy for Handling Flexible Materials" (EP/T033517/1). Z. Chen and S. Zhang contributed equally to this work. \emph{*Corresponding authors: Shan Luo and Bin Fang.}}} 
  \thanks{\final{$^{1}$Z. Chen and S. Luo are with the Department of Engineering, King’s College London, London WC2R 2LS, United Kingdom. E-mail: {\tt\footnotesize czx980730@gmail.com; shan.luo@kcl.ac.uk}.}}
 \thanks{\final{$^{2}$Z. Chen, B. Fang and  F. Sun are with the Institute for Artificial Intelligence, Department of Computer Science and Technology, Beijing National Research Center for Information Science and Technology, Tsinghua University, Beijing 100084, China. E-mail: {\tt\footnotesize czx980730@gmail.com; fangbin@tsinghua.edu.cn; fcsun@mail.tsinghua.edu.cn }.}}
 \thanks{\final{$^{3}$S. Zhang is with the School of Engineering and Technology, China University of Geosciences (Beijing), Beijing 100083, China. E-mail: {\tt\footnotesize zhangshixin@email.cugb.edu.cn}.}}
 \thanks{\final{Digital Object Identifier (DOI): see top of this page.}}
 }
 
\markboth{IEEE Robotics and Automation Letters. Preprint Version. Accepted January, 2023}
{Chen \MakeLowercase{\textit{et al.}}: Tacchi: A Pluggable \zc{and Low Computational Cost} Elastomer Deformation Simulator for Optical Tactile Sensors}


\maketitle

\begin{abstract}
Simulation is widely applied in robotics research to save time and resources. There have been several works to simulate optical tactile sensors that leverage either a smoothing method or Finite Element Method (FEM). However, elastomer deformation physics is not considered in the former method, whereas the latter requires a massive amount of computational resources like a computer cluster. In this work, we propose a pluggable and \blue{low computational cost} simulator using the Taichi programming language for simulating optical tactile sensors, named as \textit{Tacchi}. \red{It reconstructs elastomer deformation using particles, which allows deformed elastomer surfaces to be rendered into tactile images and reveals contact information \blue{without suffering from high computational costs.} Tacchi facilitates creating realistic tactile images in simulation, e.g., ones that capture wear-and-tear defects on object surfaces. In addition, the proposed Tacchi can be integrated with robotics simulators for a robot system simulation. Experiment results showed that Tacchi can produce images with better similarity to real images and achieved higher Sim2Real accuracy compared to the existing methods. Moreover, it can be connected with MuJoCo and Gazebo \blue{with only the requirement of 1G memory space in GPU compared to a computer cluster applied for FEM.}} With Tacchi, physical robot simulation with optical tactile sensors becomes possible. All the materials in this paper are available at \url{https://github.com/zixichen007115/Tacchi}.
\end{abstract}

\section{Introduction}
\IEEEPARstart{S}{imulation} plays a significant role in robotic research. \red{Experiments in the real world are costly and time-consuming while posing potential risks of wear and accidents to the robots, but simulation can provide data as preliminary experiments without such risks. Besides, data-driven methods like neural networks and deep learning are utilized in robotics for sensing\cite{DG21} and control\cite{CL19}. An extensive amount of data is required, and simulation provides an efficient way to collect data with the potential for transferring trained robot agents to the real world.}

In the past years, various simulation approaches have been proposed for object simulation, such as the Finite Element Method (FEM)~\cite{CS19} and key points~\cite{ZH19}. There have been some commonly used simulators for robotics, such as Gazebo~\cite{NK04}, Pybullet~\cite{EC16} and MuJoCo~\cite{ET12}. Although they are able to simulate robot components like robot arms and grippers, tactile sensor simulation is still a challenging task, which hinders physical robot simulation with tactile sensing.

\begin{figure}[t]
\centering
\includegraphics[width=3.4in]{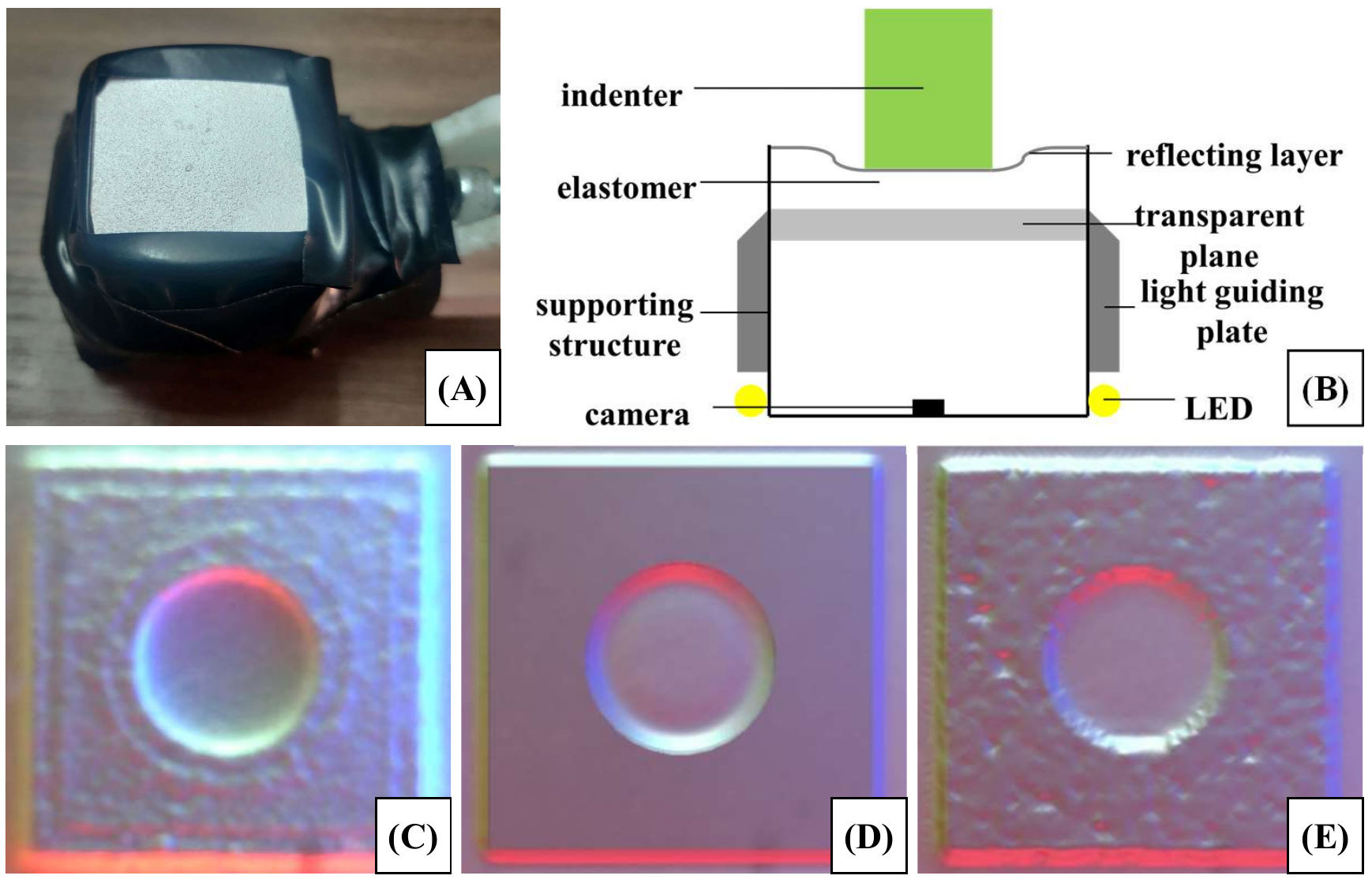}
\caption{(\textbf{A}): The GelSight sensor. (\textbf{B}): GelSight diagram. This diagram shows that the elastomer layer deforms during interaction with an indenter. (\textbf{C}), (\textbf{D}), and (\textbf{E}): Optical tactile images collected from the real sensor~\cite{WY17}, \zc{pseudo tactile image generated from ~\cite{DG21} and Tacchi proposed in this work}. As shown in the figures, Tacchi offers more realistic tactile images in simulation with surface defects.}
\label{fig1}
\end{figure}

Tactile sensing is indispensable for robot contact control, and many tactile sensors have been developed in recent years~\cite{JF12,ZS22}. Thanks to the ability to produce high-resolution tactile images, optical tactile sensors~\cite{WY17,BW18,BF18,DG20} that use cameras to capture deformation of soft elastomer have been favoured for robotics research. An example is \textit{GelSight}, as shown in Fig.~\ref{fig1}-(\textbf{A}). A GelSight sensor is composed of a camera, a set of LEDs, and a soft elastomer. During interaction with objects, the deformed opaque surface of the elastomer reflects light from the LEDs to the camera, and the tactile images captured by the camera can offer contact information. Due to their low cost and high resolution, optical tactile sensors have been widely used in real experiments~\green{\cite{luo2018vitac,cao2020spatio,SS22}}, but challenges in simulating deformations of their soft elastomer layers impede having them in simulation.

The main challenge of optical tactile stimulation is elastomer deformation reconstruction. Deformable object simulation is a complex problem caused by high degrees of freedom. Elastomer deformation is first approximated in~\cite{gomes2019gelsight} using pseudo depth maps applied with a smoothing method. However, in this method the smoothing parameters are fine-tuned only based on the optical comparison, ignoring the elastomer elasticity properties. Hence, this method can only be applied when a normal force is applied. Meanwhile, in some other works \cite{CS19}, simulation software like Abaqus is used to provide detailed elastomer deformations. However, they suffer from a high computational cost for FEM simulation. Considering the high speed of the other robot simulation processes, this time-consuming approach is unsuitable for a single sensor simulation.

To address these issues, we propose a pluggable and \zc{low computational cost} optical tactile sensor simulator \textit{Tacchi}. This simulator is named after tactile sensor simulation and the applied parallel programming language, Taichi \cite{YH19}. In Tacchi, particles are utilized to represent elastomers of the optical tactile sensors, and their motions represent elastomer deformation. Surface depth maps are generated from the particles on the elastomer surface, which are then rendered into tactile images. Experimental results show that Tacchi can be connected with robotics simulators like \zc{Gazebo and} MuJoCo. Our method produces higher quality tactile images than the existing methods~\cite{DG21, AA21} in terms of both tactile image quality and Sim2Real transfer learning. In addition, Tacchi can add random defects to simulate textures on real objects and improve the Sim2Real performance. Examples of the real and simulated tactile images from different approaches are shown in Fig. \ref{fig1}-(\textbf{C}), (\textbf{D}).

The contributions of this paper are summarized as follows:
\begin{enumerate}
\item We develop a novel simulator \textit{Tacchi} that is pluggable and \zc{requires low computation space. This is the first optical tactile sensor simulator that utilizes physical deformation simulation.} The tactile images from Tacchi achieve satisfactory similarity to the real tactile images.
\item We connect Tacchi and the commonly used simulators, \zc{Gazebo and} MuJoCo, for robot simulation. The experimental results show that this simulation can achieve robot simulation \zc{and include tactile sensors.}
\item We investigate the effect of particle number on the simulated tactile images with Tacchi, and the results show that our approach achieves superior performance in Sim2Real learning.
\end{enumerate}

The rest of the paper is structured as follows: Section \ref{sec:2} includes works related to elastomer simulation and sensor simulation; Section \ref{sec:3} introduces our proposed Tacchi simulator and its connection with \zc{Gazebo and} MuJoCo; Section \ref{sec:4} describes the experimental setup, including the real and simulation setups; Section \ref{sec:5} shows the experimental results; Section \ref{sec:6} summarizes the work.

\section{Related Work}
\label{sec:2}
\subsection{Elastomer simulation}
It is challenging to simulate elastomers like gel layers in optical tactile sensors due to their deformation properties and high degrees of freedom. FEM has been widely applied to simulate deformation in \cite{CS19,ZS22}. Nodes and facets are applied in \cite{LZ17} to represent a rubber sphere and a foam cube. Based on such a model, a strategy to grasp an elastomer with a multi-finger robot hand is introduced.

\red{In optical tactile sensors, various methods are applied for elastomer simulation. Physics-based methods, which simulate the deformation with some physical principles, have been utilized. For example, FEM is applied in \cite{CS19} for force distribution estimation. Although this work utilizes physics-based simulation, it requires massive computational resources and the simulation is achieved by using a high performance computing cluster. In FEM, each element is connected with the other ones, and deformation simulation may also lead to mesh distortion problems like irregular meshes or even negative element volume~\cite{NL93}. Meanwhile, some non-physics-based methods try to avoid physical interaction simulation with smoothing methods. Pseudo elastomer surfaces are reconstructed in \cite{DG21} by smoothing depth maps with Gaussian kernels. Similarly, \cite{AA21} uses object depths for the contacted area and kernel smoothing method for the uncontacted area. PyBullet~\cite{EC16} is utilized in \cite{SW20} as the object and sensor simulator. Smooth depth maps are generated by either pre-processing the object meshes or applying data augmentation. Although these methods simulate tactile images, they do not simulate physical deformation and only produce depth maps by smoothing, which do not have physical meaning.}

\red{To address these issues, we employ the Material Point Method (MPM) for physics-based elastomer simulation. This method applies particles to represent objects and an imaginary grid for contact simulation. The grid exchanges object information with particles during the simulation. Compared with other physics-based methods like FEM, MPM does not suffer from mesh distortion thanks to independent particles~\cite{SB04} and has a low computational cost thanks to the use of Taichi \cite{YH19}, a parallel programming language. Compared with non-physics-based methods, MPM simulates physics-based interactions and provides realistic results.}

\subsection{Sensor simulation}
Various sensors have been applied in robot systems, and sensor simulation plays an essential role before deploying them into the real world. Common sensors like cameras and force sensors have been integrated into mainstream robot simulators such as Gazebo\cite{NK04} and MuJoCo\cite{ET12}. 

In \cite{DG21}, optical tactile sensors were first simulated by smoothing depth maps collected from Gazebo. In \cite{SW20}, the meshes of both the object and the sensor are first loaded in OpenGL~\cite{DS13} and then updated during the simulation in PyBullet. \zc{In this work, depth maps can either be obtained from PyBullet or generated in OpenGL together with tactile images. However, these depth maps from widely applied robot simulators do not include physics-based deformation.}

Engineering simulation software is also utilized for tactile sensor simulation. COMSOL Multiphysics is exploited in \cite{HW19} to predict magnetic fields under different conductive target situations. In \cite{CS19}, Abaqus is used to simulate elastomer deformation. Although these software programs show excellent simulation results, most of them are packaged, and it is challenging to connect them with robot simulators due to incompatibility and time-consuming FEM algorithms. 

In this work, we develop a novel simulator \textit{Tacchi} to simulate elastomer deformation by utilizing a high-performance numerical computation language Taichi~\cite{YH19}. Thanks to its parallel programming and concise data structure, Tacchi can be run \zc{with a low requirement of computational resources}, and it is simple to connect to simulators like MuJoCo \zc{and Gazebo.}

\section{Method}
\label{sec:3}
In this section, the proposed simulator Tacchi is introduced and its connection with the robot simulator MuJoCo \zc{and Gazebo }is also detailed. Specifically, the elastomer simulation is first introduced in Subsection \ref{C.A}; a tactile image rendering method based on elastomer simulation in Tacchi is then proposed in Subsection \ref{C.B}; the integration of Tacchi with a robot simulator is detailed in Subsection \ref{C.C}; and the potential applications of Tacchi in the simulation of other sensors and robot tasks are introduced in Subsection \ref{C.D}.

\subsection{Elastomer simulation}
\label{C.A}

\begin{figure*}[t]
\centering
\includegraphics[width=6.5in]{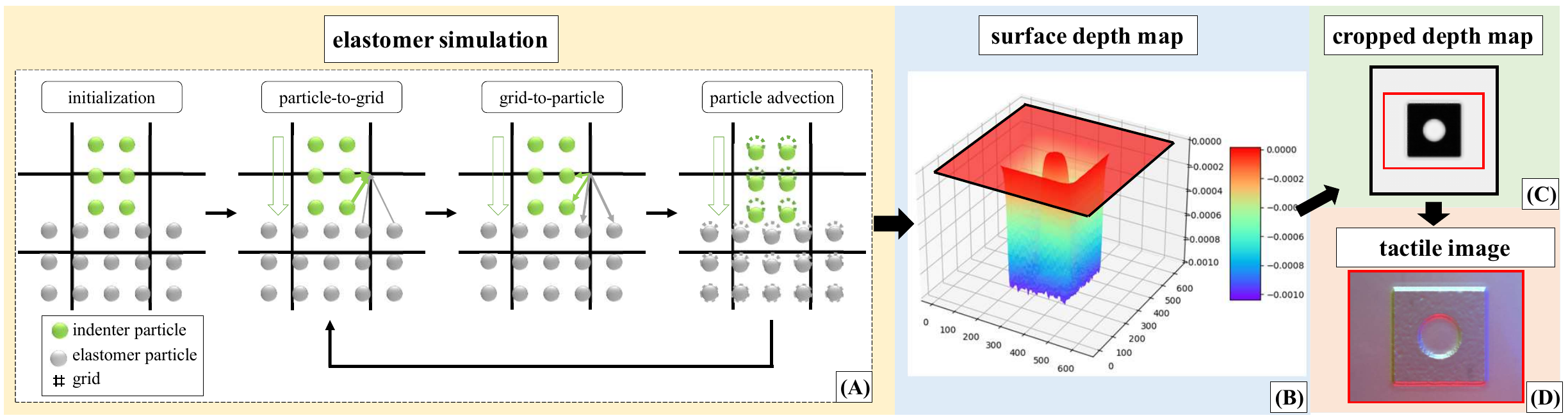}
\caption{Tactile image generation process. (\textbf{A}): Elastomer simulation. Grey particles represent the elastomer, and green particles represent the indenter. The particles and grid nodes are first initialized; the particle information is transferred to the grid in particle-to-grid; then the object information (i.e., the indenter's motion in this case) is allocated back to nearby particles by the grid nodes in grid-to-particle; finally, particles are moved according to the object information. (\textbf{B}): With the discrete surface particles, a continuous surface depth map can be reconstructed via a 2D interpolation method. (\textbf{C}): The depth map is cropped into $480\times640$ to fit the image size from the Gelsight. The cropped image is shown by the red frame. (\textbf{D}): A tactile image is generated based on the cropped depth map and the rendering method.}
\label{fig2}
\end{figure*}

{In Tacchi, we apply the Material Point Method (MPM) for elastomer simulation. MPM applies particles to represent objects and an imaginary grid for contact simulation. The grid exchanges information with particles during the simulation. FEM and MPM use meshes and particles as their elements, respectively. For both approaches, each element interacts with the nearby elements based on their geometry shapes and relative positions. For FEM, there will be mesh distortions because deformation may lead to problems like irregular meshes or even negative element volume, which results in low accuracy of stress \cite{NL93}. In contrast, for MPM, particles exchange information with an imaginary grid, and this method does not suffer from mesh distortions.} 

We first introduce the elastomer simulation in optical tactile sensors. To describe this method, we simulate an elastomer pressed by a rigid indenter as an example, as shown in Fig. \ref{fig2}-(\textbf{A}). In Tacchi, both the elastomer (grey dots in Fig. \ref{fig2}-(\textbf{A})) and the object that interacts with the elastomer, i.e., an indenter (green dots in Fig. \ref{fig2}-(\textbf{A})) in this example, are represented using particles. The object information, e.g., the position and the velocity, are stored in these particles.

In the simulation, it is necessary to decide which particles are included in the interaction between the elastomer and the object and how they interact. To this end, a fixed imaginary grid is introduced, as shown in Fig. \ref{fig2}-(\textbf{A}), so as to have the particles exchange object information via the grid nodes.

In each simulation step (i.e., the motions of the particles are updated for one time), as shown in Fig. \ref{fig2}-(\textbf{A}), the object information is first collected by the grid nodes from nearby particles (we name it as \textit{particle-to-grid}); the interaction between the particles of the elastomer and the indenter, and also the object elastic deformation, are then calculated by the grid nodes and shared with each particle (we name it as \textit{grid-to-particle}); the states of the particles will be updated accordingly afterwards (we name it as \textit{particle advection}).

\noindent {\bf{Initialization:}} Firstly, we initialize particles and grid nodes. A set of $m$ particles are used to represent the elastomer and the indenter, and a grid of  $n$ grid nodes is used to record the object information with a fixed coordinate frame. For the $p$-th particle, its position $x_p \in R^3$ and its velocity $v_p \in R^3$ are considered for the simulation of the object motions. For deformation simulation, similar to \cite{YW21}, we introduce a deformation map ${\phi}_{p}: R^3 \rightarrow R^3$ that records the initial and final position in a simulation step for each particle. The deformation gradient $F_p\in R^{3\times3}$ can be derived as

\begin{equation}
\label{eq1}
F_p = \frac{\partial \phi_p}{\partial x_p}(x_p),
\end{equation}
where $F_p$ is set as a three-dimensional identity matrix $I_{3 \times 3}$ in the initialization.

In addition, an affine velocity matrix $C_p\in R^{3\times3}$ that records the velocity of the neighboring particles \cite{YW21} is introduced to reduce information loss during information exchange between the particles and the grid nodes. It is initialized as $0^{3 \times{3}}$ since the particles are static and will be updated afterward.

\noindent{\bf{Particle-to-grid:}} In each simulation step, particle information is first transferred to the grid by calculating momentum and mass in each grid node. It can be seen as simulating part of the object surrounding the grid node by quadratic B-Spline weighting interpolation.
Specifically, the mass and momentum of the nearby particles are collected. The mass $M_i$ of the $i$-th grid node is

\begin{equation}
\label{eq2}
M_i = \sum_{j \in \mathbb{G}_i} \sum_{p \in \mathbb{P}_j} {w_{jp}} m_p,
\end{equation}
where $m_p$ denotes the mass of the $p$-th particle, $\mathbb{G}_i$ denotes $3 \times 3 \times 3$ grid nodes which contain the $i$-th grid node and its neighboring grid nodes, $\mathbb{P}_j$ denotes the particles inside the $j$-th grid, and $w_{ij}$ is the weight parameter of the $i$-th grid node and the $j$-th particle for weighting interpolation, which is calculated by quadratic B-Spline. By applying such weight parameters, close particles contribute more to $M_i$ than far ones. We follow \cite{CJ16} to have the same weight parameters for mass and the following momentum calculation.

The grid momentum ${MG}_i$ of the $i$-th grid node can be obtained by calculating the momentum resulted from the particle motion ${MM}_i$ and the momentum resulted from the elasticity ${ME}_i$:
\begin{equation}
\label{eq3}
{MG}_i = {MM}_i + {ME}_i.
\end{equation}

\noindent Here $MM_i$ is calculated by collecting the velocity and affine velocity of the nearby particles:
\begin{equation}
\label{eq4}
MM_i = \sum_{j \in \mathbb{G}_i} \sum_{p \in \mathbb{P}_j} \shan{w_{jp}} (m_p v_p + C_p(X_j - x_p)),
\end{equation}
{where $X_j$ denotes the position of the $j$-th neighboring node of the $i$-th grid node.}

$ ME_i $ can be obtained as in \cite{YW21} by: 
\begin{equation}
\label{eq5}
ME_i = - \triangle t \sum_{j \in \mathbb{G}_i} \sum_{p \in \mathbb{P}_j} \frac{4}{\triangle X^2} \shan{w_{jp}} V_p^0 S_p(X_j - x_p),
\end{equation}
where $\triangle t$ is the time interval between two adjacent steps, $\triangle X$ is the grid node interval, $V_p^0$ denotes the initial particle volume, $S_p$ is the elasticity force for the $p$-th particle \cite{YW21}.

After obtaining $ M_i $ and $ MG_i $ using Eq. \ref{eq2} and Eq. \ref{eq3} respectively, the object velocity close to the $i$-th grid node $V_i$ can be computed as:  
\begin{equation}
\label{eq6}
V_i =\frac{MG_i}{M_i}.
\end{equation}

\noindent {\bf{Grid-to-particles:}} The states of the particles are then updated with the previous states of the grid nodes and the particles. It can be seen as simulating part of the object surrounding the particle by quadratic B-Spline weighting interpolation. Suppose that the states of the $k$-th step, i.e., the velocity $v_p^{(k)}$, the affine velocity $C_p^{(k)}$, and the deformation gradient $F_p^{(k)}$, are known, $v_p^{(k+1)}$, $C_p^{(k+1)}$ and $F_p^{(k+1)}$ in the $k+1$ step can be obtained as

\begin{equation}
\label{eq7}
v_p^{(k+1)} = \sum_{i \in \mathbb{G'}_p} w_{ip}V_i^{(k)},
\end{equation}
\begin{equation}
\label{eq8}
C_p^{(k+1)} = \frac{4}{\triangle X^2} \sum_{i \in \mathbb{G'}_p} w_{ip} v_p^{(k+1)}(X_i - x_p^{(k)}),
\end{equation}
\begin{equation}
\label{eq9}
F_p^{(k+1)} = (I + \triangle t C_p^{(k+1)}) F_p^{(k)},
\end{equation}
where $\mathbb{G'}_p$ represents the $3 \times 3 \times 3$ grid nodes surrounding the $p$-th particle. 

\noindent {\bf{Particle advection:}} \shan{With Eq. \ref{eq8} and Eq. \ref{eq9}, the states of particles can be updated. However, there are two special cases: the velocities of the particles on the bottom layers of the elastomer are set as $0$ since they are fixed onto the sensor (and the sensor is static in this example); as the indenter is rigid, the particles representing the indenter share the same velocity, i.e., the velocity of the indenter. As a result, $v_p^{(k+1)}$ for these special particles are obtained as:}

\begin{equation}
\label{eq10}
v_p^{(k+1)} =
\begin{cases}
v_i, & p \in \mathbb{I},\\
0,   & p \in \mathbb{B},\\
\end{cases}
\end{equation}
where $v_i$ denotes the velocity of the indenter; $\mathbb{I}$ denotes particles representing the indenter; $\mathbb{B}$ denotes the particles on the bottom layers of the elastomer.

Finally, the particles move with the above computed velocities and the position of each particle $x_p^{(k+1)}$ in the $k+1$ step is
\begin{equation}
\label{eq11}
x_p^{(k+1)} = x_p^{(k)} + \triangle t v_p^{(k+1)}.
\end{equation}

\subsection{Tactile image rendering}
\label{C.B}

As shown in Fig. \ref{fig2}-(\textbf{B}), (\textbf{C}), and (\textbf{D}), after particle motion surface particles and the Phong’s model \cite{BP75} are used to generate the tactile images. Although there may be various methods to produce tactile images from the particles, we follow \cite{DG21}'s method by generating depth maps first and rendering them. A 2D interpolation method is applied to provide a continuous depth map based on discrete particles representing the elastomer surface, as shown in Fig. \ref{fig2}-(\textbf{B}). It should be highlighted that the obtained depth map here is from the physics-based elastomer deformation, instead of the depth map of the indenter like \cite{DG21}. In this way, the resulted tactile image will be more realistic as in the real sensor the camera also captures the elastomer deformation. 

As shown in Fig. \ref{fig2}-(\textbf{B}), the depth map of the whole square elastomer surface is provided in Tacchi. However, due to its limited field of view, the camera in the Gelsight~\cite{WY17} sensor generates images with a resolution of $480\times640$, in which only a partial area of the sensor is captured instead of the whole surface. To this end, only this partial area of the depth map will be rendered into tactile images (Fig. \ref{fig2}-(\textbf{C})). In addition, there are misalignments between the desired and real setups, due to the calibration errors of the camera and the indentor's movements. In \cite{DG21}, some cropping and alignment approaches are proposed to match real and the simulated tactile images so as to compensate the misalignments. We follow the \textit{Alignment (per object)} approach \cite{DG21}, with the cropped depth map shown in Fig. \ref{fig2}-(\textbf{C}).

We follow \cite{DG21} to simulate illumination of the GelSight sensor using the Phong’s model \cite{BP75}. With the obtained depth map $D$ and light resources $L$, the tactile image $I$ can be obtained, as shown in Fig.~\ref{fig2}-(\textbf{D}):

\begin{equation}
\label{eq12}
I = k_a i_a + \sum_{m \in L}(k_d (\widehat{L}_m \cdot \widehat{N})i_{m,d} + k_s (\widehat{R}_m \cdot \widehat{V})^{\alpha} i_{m,s}),
\end{equation}
\begin{equation}
\label{eq13}
\widehat{R}_m = 2(\widehat{L}_m \cdot \widehat{N})\widehat{N}-\widehat{L}_m,
\end{equation}
\begin{equation}
\begin{aligned}
\label{eq14}
 \widehat{N} &= <\frac{\partial H}{\partial x},\frac{\partial H}{\partial y},-1>\\
 &= <\frac{H}{2r} * [-1, 0, 1],\frac{H}{2r} * [-1, 0, 1]^T,-1>,
\end{aligned}
\end{equation}
where $k_a$, $k_d$, $k_s$, and $\alpha$ are surface reflectance property parameters; $r$ is the pixel-to-meter ratio; $i_a$ is the background light that is not from the light resources; $i_{m,d}$ and $i_{m,s}$ are the intensities of the diffuse and specular reflections of light source $m$ respectively; $\widehat{L}_m$ and $\widehat{R}_m$ are emission and reflection direction of the light resource $m$; $\widehat{V}$ is the direction from the surface point to camera; $\widehat{N}$ is the elastomer surface normal; $*$ stands for 2D convolution operation in this equation. These parameters for simulating the GelSight are detailed in \cite{DG21}.

\subsection{Connection with a robot simulator}
\label{C.C}

\begin{figure*}[t]
\centering
\includegraphics[width=6.2in]{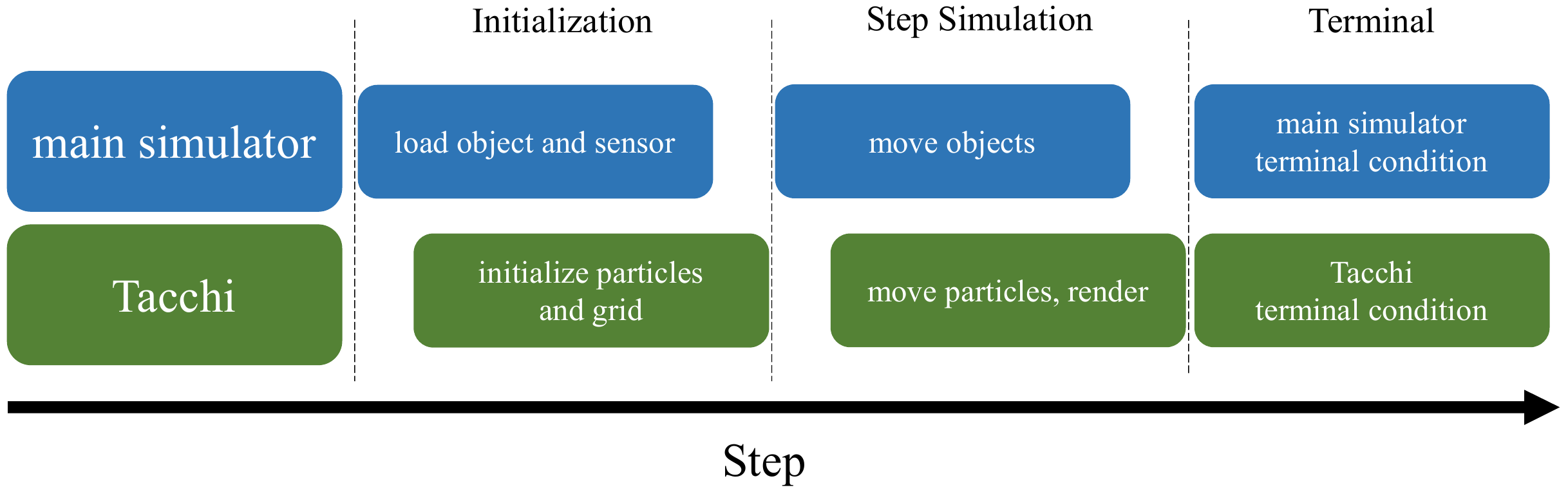}
\caption{Connection with a robot simulator. The misalignment denotes the time sequence in each simulation step. In initialization, the robot simulator loads models as the main simulator, and Tacchi initializes particles and a grid. During the simulation, the main simulator moves the objects based on the desired policy, and Tacchi obtains the velocity of the objects from the main simulator and moves the particles. Tactile images are rendered from the depth maps. Terminal conditions are proposed in the main simulator and Tacchi.}
\label{fig3}
\end{figure*}

This subsection introduces the connection of Tacci with a robot simulator. As shown in Fig. \ref{fig3}, Tacchi generates tactile images, and the robot simulator is applied as the main simulator to simulate the motions of the objects and the robot, etc. \zc{MuJoCo and Gazebo are} used as the main simulator in this paper. However, any robot simulator that can provide the real-time \zc{position or }velocity of the sensor and the contacted objects can also be used. First, the object and sensor models, usually mesh files, are loaded in \zc{the main simulator}, and they are placed in the initial positions. As Tacchi requires particle information, point clouds from the object and sensor models are used to initialize the particles of the object and the sensor.

In every simulation step, the object, i.e., the indenter in this example, moves according to the control policy in \zc{the main simulator}. The object position or velocity is acquired by Tacchi from the main simulator to simulate the object's motion. Then depth maps are generated from the elastomer deformation and are rendered into tactile images. During the simulation, if a terminal condition is met, the simulation will terminate. For example, if the indenter reaches the desired depth in \zc{the main simulator} or Tacchi, the simulation will end.

\subsection{Application of Tacchi to other sensors and tasks}
\label{C.D}
This subsection describes how to apply Tacchi to other tasks like robot grasping, and other optical tactile sensors~\cite{BF18,DG20,gomes2020blocks} or soft inductive tactile sensors \cite{HW19}. As shown in the above sections, only the elastomer in the tactile sensor and the objects which may contact the sensor are involved in Tacchi. To this end, there are only a few points to consider when using Tacchi to simulate other sensors or in other tasks: 
\begin{itemize}
    \item In initialization, as the particles are desired to be inside the grid during the simulation, the grid needs to be a little larger than the particle range; 
    \item The grid node density should strike a balance between simulation quality and speed;
    \item As the particle density will affect object surface roughness and computation cost, it needs to be determined according to the simulation demand; 
    \item \red{For a new object or elastomer, its point clouds can be used, and the particles can be aligned according to the point positions in the point clouds. Elasticity parameters like Young's modulus and Poisson's ratio can be initially decided by measuring the applied material or using their standard parameters, and then fine-tuned based on the sensor outputs.}
\end{itemize}

\section{Experimental Setup}
\label{sec:4}

To evaluate Tacchi, we collect tactile images in the real world and simulation. Subsection \ref{D.A} introduces real experimental devices and data collection. Subsection \ref{D.B} includes the simulation setup like Tacchi parameters, indentor loading, and \zc{main simulator} setup.

\subsection{Real World setup}
\label{D.A}

\begin{figure}[ht]
\centering
\includegraphics[width=3.3in]{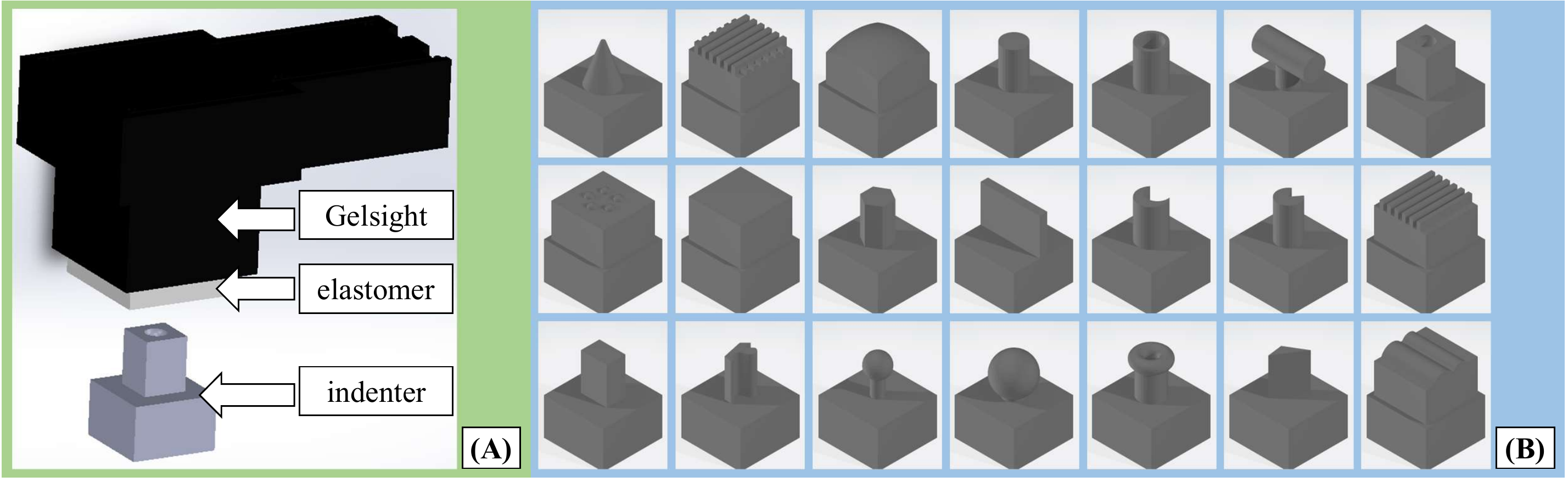}
\caption{(\textbf{A}): Real experiment diagram. The GelSight sensor presses against the indenter. (\textbf{B}): Twenty one indenters included in this work. The first row: {\tt $cone$}, {\tt $cross\_lines$}, {\tt $curved\_surface$}, {\tt $cylinder$}, {\tt $cylinder\_shell$}, {\tt $cylinder\_side$}, {\tt $dot\_in$}. The second row: {\tt $dots$}, {\tt $flat\_slab$}, {\tt $hexagon$}, {\tt $line$}, {\tt $moon$}, {\tt $pacman$}, {\tt $parallel\_lines$}. The third row: {\tt $prism$}, {\tt $random$}, {\tt $sphere$}, {\tt $sphere2$}, {\tt $torus$}, {\tt $triangle$}, {\tt $wave1$}.}
\label{fig4}
\end{figure}

We adopt the dataset from \cite{DG21} in this work. In this case, GelSight and the objects in \cite{DG21} are included in the real world setup. In the experiment, a Gelsight is pressed against an indenter, as shown in Fig. \ref{fig4}-(\textbf{A}). The dataset contains 21 objects with various shapes as indenters, which are shown in Fig. \ref{fig4}-(\textbf{B}). $3 \times 3$ positions are chosen for press experiments, and the horizontal step is $1$mm. The indention depth improves from $0$ to $1$mm, and the vertical step is $0.1$mm, as a result $11$ images are collected at each position. Considering $21$ indenters, this experiment results in $3 \times 3 \times 11 \times 21 = 2,079$ tactile images. More details can be found in \cite{DG21}.

\subsection{Virtual World setup}
\label{D.B}

\begin{figure}[t]
\centering
\includegraphics[width=3.3in]{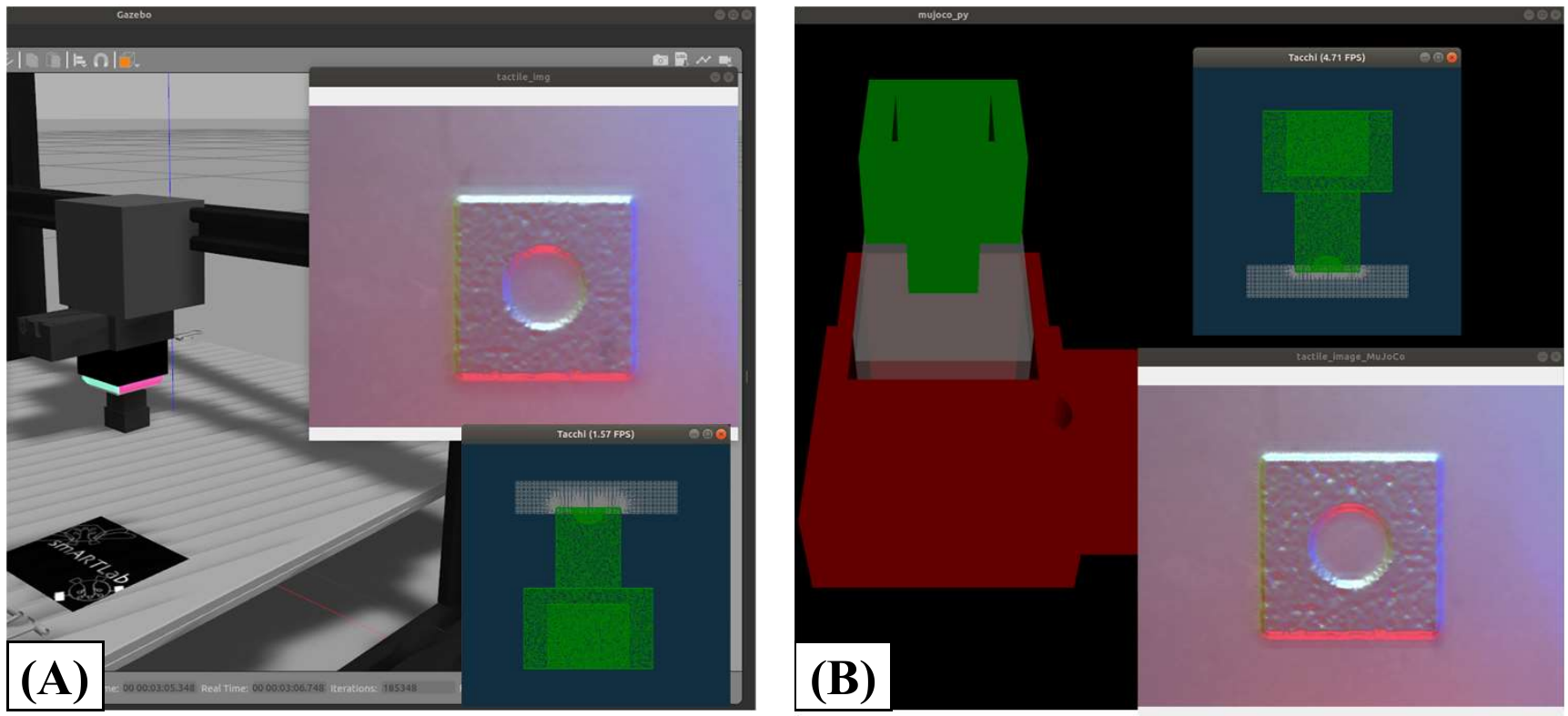}
\caption{ \zc{(\textbf{A}): Tacchi connected with Gazebo. The sensor moves in Gazebo, and tactile images are generated by Tacchi. (\textbf{B}): Tacchi connected with MuJoCo. The indenter is pressed against the elastomer, and tactile images are generated by Tacchi.}}
\label{fig5}
\end{figure}

\zc{Fig. \ref{fig5}-(\textbf{A}) and Fig. \ref{fig5}-(\textbf{B}) show the Tacchi simulator connected with Gazebo and MuJoCo, respectively.} In the Tacchi interface, an elastomer layer (grey dots) and an indenter (green dots) are included in the virtual world. In our work, the GelSight elastomer is composed of $101 \times 101 \times 21$ particles since the shape of the elastomer layer in the real GelSight sensor is $20mm \times 20mm \times 4mm$. Young's modulus and Poisson's ratio are $1.45\times10^5$ and $0.45$, and these parameters are decided according to \cite{WY17} and some preliminary experiments. The grid contains $256 \times 256 \times 256$ grid nodes, and the grid edge is $33$mm. More grid nodes lead to better simulation quality but slow down the simulation process. In general, all particles should be inside the grid during the simulation, and the grid interval should be larger than the particle interval. Based on such principles and after some preliminary experiments on different grid node densities, we choose such grid parameters. Although we only apply GelSight as the optical tactile sensor in this work, Tacchi can be utilized to simulate other tactile sensors by initializing particles with corresponding shapes.

To create various indenters in Tacchi, the point positions in the 3D point clouds are used to initialize the indenter particles in Tacchi. If only 3D mesh files are available, some software programs and libraries like {\tt Open3D} and {\tt Point Cloud Library} can be utilized to extract point clouds from mesh files. Though all the points in the point clouds can be included, we change the particle density to accommodate the computational requirements. Also, the intervals among particles can simulate defects on the object surfaces. This feature will be exploited in Section \ref{sec:5}.

\blue{Gazebo and }MuJoCo simulation interface are shown in Fig. \ref{fig5}-(\textbf{A}) and Fig. \ref{fig5}-(\textbf{B}). In the main simulator, we simulate the sensor and indenter by loading mesh files. \blue{Gazebo provides the indenter position, and Tacchi generates the corresponding tactile image.} There is an interface in MuJoCo that can provide the indenter velocity, and the indenter in Tacchi shares the same velocity with that in MuJoCo by obtaining velocity through this interface. The indenter velocity control in Tacchi is achieved by Eq. \ref{eq11}. Although we only include \blue{Gazebo and} MuJoCo in this work, any robot simulator which can provide the real-time indenter position or velocity has the potential to be the main simulator with Tacchi.

\section{Experimental Results}
\label{sec:5}

This section shows the experimental results of applying Tacchi as the optical tactile sensor simulator. We first introduce \zc{the computation cost} in Subsection \ref{E.A}. To evaluate our approach, Subsection \ref{E.B} compares the pseudo tactile images from \cite{DG21, AA21} and Tacchi. We also show images with different particle numbers to explore the effect of this parameter. Sim2Real learning is applied for evaluation in Subsection \ref{E.C}. 

\subsection{Computational Cost}
\label{E.A}

\red{High computational cost is the main problem of physics-based simulation. For example, the FEM simulation is achieved with high computational costs using a high performance computing cluster in \cite{CS19}. However, the parallel programming language Taichi\cite{YH19} highly decreases the demand for computational resources. To simulate the sensor interaction with an intended in this work, over $10^6$ particles and $10^7$ grid nodes are simulated, but it only requires 1G memory in a GTX 1050Ti graphics card. Compared with FEM in \cite{CS19}, Tacchi requires much lower computational cost.}

\subsection{Pseudo image quality}
\label{E.B}

\begin{figure*}[t]
\centering
\includegraphics[width=6.2in]{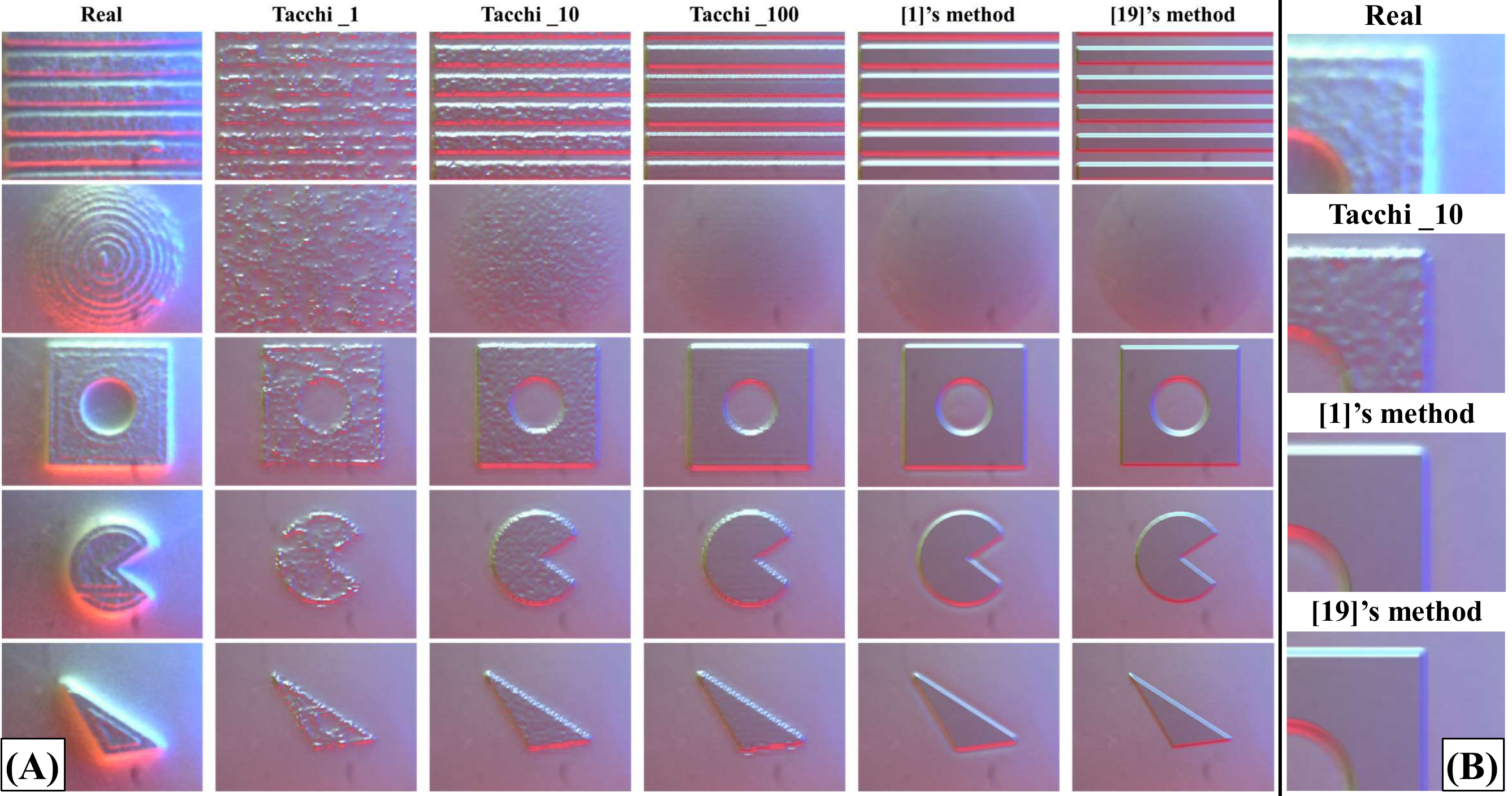}
\caption{\zc{(\textbf{A}): Tactile images from real dataset in \cite{DG21} (top column), Tacchi$\_1$ (2nd column), Tacchi$\_10$ (3rd column), Tacchi$\_100$ (4th column), \cite{DG21}’s method (5th column),  \cite{AA21}’s method (6th column). These rows show the tactile images of {\tt $cross\_lines$}, {\tt $curved\_surfaces$}, {\tt $dot\_in$}, {\tt $pacman$}, and {\tt $triangle$} respectively. Tacchi$\_100$, \cite{AA21} and \cite{DG21} show similar images but have slight differences caused by surface defects. Tacchi$\_10$ provides random defects on the object surface, and Tacchi$\_1$ generates too many defects that harm object shapes. (\textbf{B}): Enlarged areas of {\tt $dot\_in$} tactile image from real dataset, Tacchi$\_10$, \cite{DG21}, and \cite{AA21}.}}
\label{fig6}
\end{figure*}

In this subsection, we evaluate the simulation methods in \zc{\cite{DG21,AA21}} and Tacchi with three metrics: Structural Similarity (SSIM), Peak Signal-to-Noise Ratio (PSNR), and Mean Absolute Error (MAE). These metrics are widely applied to evaluate image similarity, and the similarity between real and generated images can be taken as pseudo image quality. \zc{Depth maps are generated based on the methods in \cite{DG21,AA21} and Tacchi, and the rendering method in \cite{DG21} is applied for deformation simulation comparison.} For Tacchi, we build the indenter models with \zc{$1\times10^4$, $10\times10^4$, and $100\times10^4$ particles and name them Tacchi$\_$1, Tacchi$\_$10, and Tacchi$\_$100}, respectively. We apply such particle numbers since preliminary experiments show that they provide indenters with excessive, appropriate, and few defects. Real tactile images, images from \zc{Tacchi$\_$1, Tacchi$\_$10, Tacchi$\_$100, \cite{DG21} and \cite{AA21}} are shown in Fig. \ref{fig6}-(\textbf{A}), and some images are enlarged for comparison in Fig. \ref{fig6}-(\textbf{B}). Objects contain defects and textures on their surfaces, and the real images show this feature by rough surfaces in Fig. \ref{fig6}-(\textbf{B}). The surfaces of images from \cite{DG21} and \cite{AA21} are smooth, while Tacchi$\_$10 produces rough surfaces similar to the real images. Moreover, the indention edges of the other simulation methods, \cite{DG21} and \cite{AA21}, are straight and clear, but the real edges are meandering and blurry, similar to images from Tacchi$\_$10.

\begin{table}[ht]
\caption{image quality comparison}
\centering
\begin{tabular}{l|c c c}
& SSIM $\uparrow$ & PSNR $\uparrow$ & MAE $\downarrow$\\
\hline
\red{Tacchi$\_$1}           & \red{$0.819\pm0.011$}    & \red{$18.43\pm3.07$}    & \red{$8.73\pm0.03\%$}\\
\red{Tacchi$\_$10}          & \red{$0.845\pm0.008$}    & \red{$\bf{18.57\pm2.89}$} & \red{$\bf{8.57\pm0.03\%}$}\\
\red{Tacchi$\_$100}         & \red{$\bf{0.857\pm0.006}$} & \red{$18.46\pm2.95$}    & \red{$8.67\pm0.03\%$}\\
\red{\cite{DG21}'s method}  & \red{$0.852\pm0.006$}    & \red{$18.51\pm3.26$}    & \red{$8.64\pm0.04\%$}\\
\red{\cite{AA21}'s method}  & \red{{$0.854\pm0.005$}}    & \red{{$18.50\pm3.36$}}    & \red{{$8.66\pm0.04\%$}}\\
\end{tabular}
\label{table1}
\end{table}

\zc{In Table \ref{table1}, Tacchi$\_$100 obtains the highest SSIM, and Tacchi$\_$10 has the best performance on the other two metrics. This discrepancy can also be seen in \cite{DG21}.} Compared with Tacchi$\_100$, Tacchi$\_$10 includes part of points in the point clouds and produces rough surfaces similar to real objects. In the real world, scratches and defects are common on object surfaces, and this feature can improve the performance of algorithms. But too many defects can harm the object shapes and lead to low image quality like Tacchi$\_$1. Our methods obtain better performance than \cite{DG21} and \cite{AA21} since their depth maps are based on kernel smoothing and ignore physical deformation, but real sensors provide depth maps by capturing the elastomer deformation, similar to our method.

\subsection{Sim2Real for object classification}
\label{E.C}

In this section, we study how Tacchi performs in a Sim2Real task. Simulated data is desired to approximate the real data, and it can be evaluated by comparing the performance of real and generated data on the same task. In this work, we apply object recognition as the evaluation task and introduce a neural network for the recognition task. To this end, we train the ResNet-18 \cite{KH16} network with an early stopping strategy, and each tactile image is cropped to $440\times440$ and resized to $256\times256$ before being fed into the network. We randomly choose $20$ images for each indenter as the test dataset and $10$ images as the validation dataset, considering that each indenter produces $3\times3\times11=99$ images. Images with the indention depth $0$mm are discarded since the indenters do not press into the elastomer in these images, and the images cannot provide object information. The training batch is $32$, and we train the network with real images, images generated by \zc{ Tacchi$\_$1, Tacchi$\_$10, Tacchi$\_$100, \cite{DG21} and \cite{AA21}} for at most $2000$ iterations. Then we test them on the test dataset of real images. The classification accuracies are shown in Table \ref{table2}.

\begin{table}[!ht]
\caption{Sim2Real Accuracy}
\centering
\begin{tabular}{l|l }
& Test Accuracy \\
\hline
\red{Real2Real}                       & \red{$99.67\pm0.23\%$}\\
\hline
\red{Sim2Real (Tacchi$\_$1)}          & \red{$51.26\pm5.91\%$}\\
\red{Sim2Real (Tacchi$\_$10)}         & \red{$\bf{61.87\pm7.18\%}$}\\
\red{Sim2Real (Tacchi$\_$100)}        & \red{$49.06\pm8.18\%$}\\
\red{Sim2Real (\cite{DG21}'s method)} & \red{$44.17\pm5.21 \%$}\\
\red{Sim2Real (\cite{AA21}'s method)} & \red{$48.19\pm6.89\%$}\\
\end{tabular}
\label{table2}
\end{table}

In Table \ref{table2}, Tacchi$\_$10 achieves the highest accuracy thanks to the realistic rough surfaces. The tactile images of Tacchi$\_$1 are the most coarse ones and acquire the lowest accuracy. Simulation methods based on ideal object models tend to provide results with few scratches and defects, such as \cite{DG21,AA21} and Tacchi$\_$100. However, most objects in the real world contain random defects and scratches. Therefore, ideal simulated models cannot be the same as real objects, and a proper number of defects can improve the simulation performance, as shown in Tacchi$\_$10, but too many defects affect the object information and harm the simulation like Tacchi$\_$1. 

\section{CONCLUSION AND DISCUSSION}
\label{sec:6}
We introduce Tacchi, a pluggable and \zc{low computational cost} elastomer deformation simulator for optical tactile sensors. It simulates elastomer deformation and produces tactile images with the depth maps generated from the deformed elastomer. Tacchi can be connected with a robot simulator and simulates \zc{with a low demand of computation resources.} The experimental results show that Tacchi provides realistic tactile images and achieves high accuracy in Sim2Real tasks, and defects on the object surfaces can be simulated by changing the number of particles. Overall, Tacchi is an effective optical tactile sensor simulator that can be applied in robot simulation. 

\finall{Only 3D printed objects are applied in this paper, and future works will explore the relationship between real object roughness and the particle sampling ratio in the simulation with different cases of materials and surfaces.}
Some optical tactile sensors contain markers on the inner elastomer surfaces which can be represented by the particles sharing the same position. By recording surface particle motion, we can simulate marker motion. We will also consider simulation of the surface frictions during sliding and rotation motions in future works.

\bibliographystyle{IEEEtran}
\bibliography{IEEEabrv,references}
\vfill
\end{document}